\documentclass[letterpaper, 10 pt, conference]{ieeeconf} 
\IEEEoverridecommandlockouts                              % This command is only needed if 
\usepackage{graphics} % for pdf, bitmapped graphics files
\usepackage{epsfig} % for postscript graphics files
\usepackage{times} % assumes new font selection scheme installed
\usepackage{amsmath} % assumes amsmath package installed
\usepackage{amssymb}  % assumes amsmath package installed
\usepackage{color}
\usepackage{comment}
\usepackage{algorithmic, algorithm} % for algorithm
\usepackage{fourier} % for arc symbol over letters
\usepackage{dblfloatfix} % for figure on the bottom of a page
\usepackage{soul}

\def\beq{\begin{equation*}}
\def\eeq{\end{equation*}}
\def\bql{\begin{equation}}
\def\eql{\end{equation}}
\def\bqn{\begin{eqnarray*}}
\def\eqn{\end{eqnarray*}}
\def\bnl{\begin{eqnarray}}
\def\enl{\end{eqnarray}}
\def\bna{\bql\begin{array}{rcl}}
\def\ena{\end{array}\eql}
\def\bnn{\beq\begin{array}{rcl}}
\def\enn{\end{array}\eeq}
\def\bma{\begin{bmatrix}}
\def\ema{\end{bmatrix}}
\def\bmx{\begin{matrix}}
\def\emx{\end{matrix}}
\def\ben{\begin{enumerate}}
\def\een{\end{enumerate}}
\def\bit{\begin{itemize}}
\def\eit{\end{itemize}}
\def\bei{\begin{itemize}}
\def\eei{\end{itemize}}
\def\bet{\begin{tabular}}
\def\eet{\end{tabular}}

\newcommand{\allcaps}[1]{\uppercase\expandafter{#1}}

\def\bfs{\begin{footnotesize}}
\def\efs{\end{footnotesize}}
\def\bss{\begin{small}}
\def\ess{\end{small}}

\title{\LARGE \bf
Infrastructure Node-based Vehicle Localization for Autonomous Driving 
}

\author{Elijah S. Lee$^{1}$, Ankit Vora$^{2}$, Armin Parchami$^{2}$, Punarjay Chakravarty$^{2}$, Gaurav Pandey$^{2}$, and Vijay Kumar$^{1}$% <-this % stops a space

\thanks{$^{1}$The authors are with the GRASP Lab, University of Pennsylvania, Philadelphia, PA 19104, USA. 
{\tt\footnotesize \{elslee, kumar\}@seas.upenn.edu}
}%
\thanks{$^{2}$The authors are with Ford Autonomous Vehicles LLC, Dearborn, MI 48124, USA. {\tt\footnotesize \{avora3, mparcham, pchakra5, gpandey2\}@ford.com}
}%
}

\begin{document}
%\onecolumn

\maketitle
\thispagestyle{empty}
\pagestyle{empty}

%%%%%%%%%%%%%%%%%%%%%%%%%%%%%%%%%%%%%%%%%%%%%%%%%%%%%%%%%%%%%%%%%%%%%%%%%%%%%%%%
\begin{abstract}
Vehicle localization is essential for autonomous vehicle (AV) navigation and Advanced Driver Assistance Systems (ADAS). Accurate vehicle localization is often achieved via expensive inertial navigation systems or by employing compute-intensive vision processing (LiDAR/camera) to augment the low-cost and noisy inertial sensors. Here we have developed a framework for fusing the information obtained from a smart infrastructure node (ix-node) with the autonomous vehicles on-board localization engine to estimate the robust and accurate pose of the ego-vehicle even with cheap inertial sensors. A smart ix-node is typically used to augment the perception capability of an autonomous vehicle, especially when the onboard perception sensors of AVs are blocked by the dynamic and static objects in the environment thereby making them ineffectual. In this work, we utilize this perception output from an ix-node to increase the localization accuracy of the AV. The fusion of ix-node perception output with the vehicle’s low-cost inertial sensors allows us to perform reliable vehicle localization without the need for relying on expensive inertial navigation systems or compute-intensive vision processing onboard the AVs. The proposed approach has been tested on real-world datasets collected from a test track in Ann Arbor, Michigan. Detailed analysis of the experimental results shows that incorporating ix-node data improves localization performance.
\end{abstract}

\section{Introduction \label{sec:intro}}

Precise vehicle localization is critical to intelligent vehicle systems and aims to minimize traffic accidents, congestion, and overall costs of road traffic. For safety, highway operation and local city roads require reliable knowledge of vehicle coordination \cite{reid2019localization}. Further, accurate positioning enables the driving assistance modules to predict future trajectory of the vehicle and thus supports Advanced Driver Assistance Systems (ADAS) such as Adaptive cruise control (ACC), Forward collision warning (FCW), and Lane departure warning system (LDW). 

\begin{figure}[t]
\centering
\includegraphics[width=.49\textwidth]
{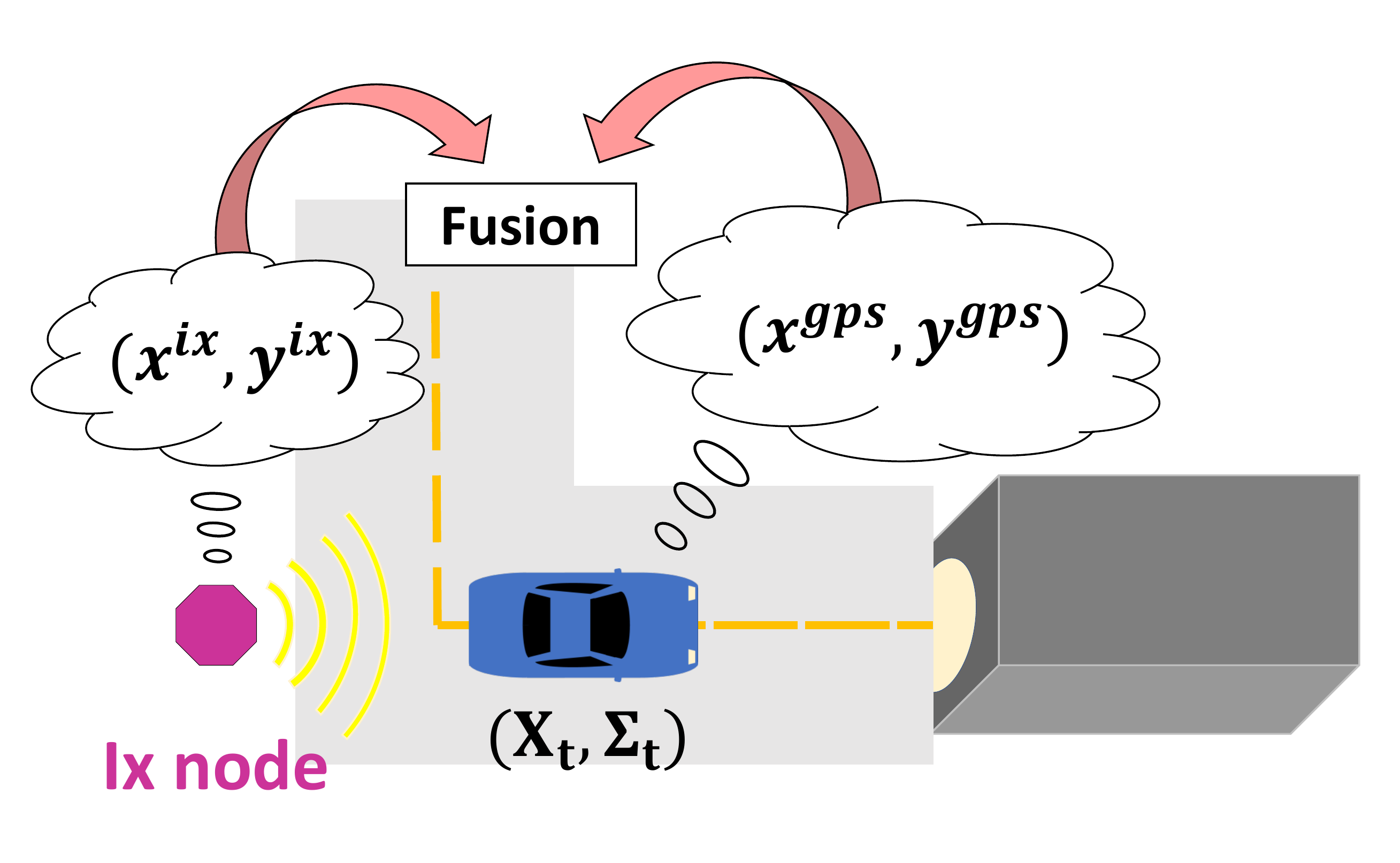}
\caption{
Ix-based vehicle localization. We fuse the ix-node perception output with the vehicle's sensor data to achieve robust and accurate vehicle localization.
}
\label{fig:intro}
\end{figure}

In this work, we propose infrastructure node-based vehicle localization to estimate vehicle coordination for autonomous vehicles. Infrastructure node \cite{ixpatent, ixmedium}, as we call ix-node, is an infrastructure based smart sensing system built to perceive the nearby vehicles and pedestrians. Similar to autonomous vehicles carrying their sensors to perceive the surrounding, ix-node has its own sensor heads to run perception algorithm and detect passing by objects. As shown in Fig. \ref{fig:intro}, we fuse the information collected from the ix-node and autonomous vehicles' sensing system to improve the overall localization performance of these vehicles. We use Extended Kalman Filter (EKF) based multisensor fusion that combines GPS and IMU data from the vehicle as well as the ix-node perception output from camera and LiDAR. To the best of our knowledge, this is the first work that performs infrastructure node-based localization with real-world testing and validation.

% Figure

The smart ix-node is a stationary infrastructure based sensing solution which provides some unique benefits to the robotics industry as listed below:
\begin{itemize}
    \item These systems have their own sensing suite and perception which can be communicated to and used by nearby passing AVs. This forms a unique sensor fusion based approach where we can assimilate information from two different view points and increase the total accuracy of the system as a whole. 
    
    \item Depending on the need at a particular road intersection, these ix-nodes can be made sensor heavy or light. This kind of flexibility results in a cost-effective autonomy solution because now, the AVs operating in that region can reduce some of their sensors, to become lightweight, and instead use the rich information from these nodes.
    
    \item These nodes increase the effective sensing field of view of AVs. If placed at strategic locations, these can fill in blind spots and reduce the probability of accidents and near miss incidents. In addition, these can stand in as a proxy for vehicle sensors just in case a failure occurs.
\end{itemize}

A number of works \cite{challita2009application,rife2011collaborative,bento2012inter} have implemented vehicle-to-vehicle (V2V) communication-based platforms where the sensor measurements depend on relative localization between those vehicles. Since ix-node is static, it can represent information in a global reference frame with higher accuracy. As a result, we use it for the purpose of localization in our work. This work utilizes the aforementioned benefits and implements ix-based vehicle localization. In short, our main contributions are:
\begin{itemize}
    \item Realization of ix-based vehicle localization. Our proposed method was tested on the real-world environment where we built our infrastructure node with sensor heads. The ix-node collects perception data of autonomous vehicles driving nearby the node.
    
    \item Improvement of localization performance by incorporating ix-node perception output with vehicle's sensor data. We use Extended Kalman Filter (EKF) based multisensor fusion that combines GPS and IMU data from the vehicle as well as the ix-node perception output from camera and LiDAR.
\end{itemize}
\section{Related Work \label{sec:related}}
Vehicle localization has been extensively studied in the past \cite{reid2019localization, bonnifait2001data, el2005road, levinson2010robust, lee2017feature, lee2019bird, localizationtechniquepatent, vora2020aerial}. The traditional methods \cite{bonnifait2001data, el2005road} integrate GPS with odometry. Wei et al. \cite{wei2011intelligent} fuse GPS with visual odometry using EKF for vehicle localization in urban environments. Although combining GPS with odometry can improve the coordination accuracy, these methods provide degraded results when sensors are noisy or sensitive to the environments. Visual odometry is feature dependent and assumes the surrounding with relevant features. Therefore, fusion of only GPS and odometry is unsuitable for reliable localization.

To achieve robust vehicle localization, LiDAR-based localization has been also explored. Kummerle et al. \cite{kummerle2009autonomous} use multi-level surface maps to localize the vehicle using particle filter. Levinson et al. \cite{levinson2010robust} address LiDAR intensity-based localization approach where the intensity offers more cues than the traditional 3D shape matching provides. Recently, Wan et al. \cite{wan2018robust} combine both LiDAR intensity and altitude information to demonstrate more robust system that can adapt to environmental changes. 

Further, learning-based methods are discussed using LiDAR. Wolcott et al. \cite{wolcott2017robust} use a Gaussian mixture map that captures LiDAR intensity and the altitude, and resulting system performs well under challenging weather conditions. Schaupp et al. \cite{schaupp2019oreos} utilize deep learning for extracting a point cloud descriptor that improves localization results. Lu et al. \cite{lu2019deepvcp} perform end-to-end point cloud registration for better localization. All of these LiDAR-based approach requires the vehicle to be equipped with the expensive LiDAR sensor and perform computationaly intensive point cloud registration.% but ix-node can offer the same benefit by sharing rich information with nearby vehicles.

Map matching based vehicle localization has been also popular in recent years. Li et al. \cite{li2018map} propose a particle filter based map matching algorithm to compute likelihood of the matched road with the topology of the lane-level map. Asghar et al. \cite{asghar2020vehicle} also execute lane level map matching and perform topological map matching using GPS, odometry, and IMU data. Vora et al. \cite{vora2020aerial} tackle the problem of reliance on prior maps by using freely available aerial imagery to match it against the live LiDAR intensity. With prior knowledge of map, these map matching based methods result in accurate localization, but ICP based matching assumes good initialization and the algorithm depends on the map.

To remedy the aforementioned problems, vehicle-to-vehicle (V2V) communication-based localization has been  studied. The V2V communication is robust to GPS signal outage and cooperative sensing capability will add reliability to localization accuracy. Challita et al. \cite{challita2009application} use V2V communication with GPS and vision-based ranging system for localizing ego vehicle. Rife et al. \cite{rife2011collaborative} utilize V2V, GPS, and vision-based lane-boundary detection for accurate vehicle positioning. Bento et al. \cite{bento2012inter} achieve V2V-based sensor fusion using encoders and inter-vehicle absolute positioning data from magnet detection. Although V2V approach performs better than GPS and odometry fusion, its robustness is supported by relative measurements. In the case of lane boundary detection method, the system is also sensitive to proper lane marking.

Since V2V-based approach depends on relative measurements, infrastructure-based localization has been recently introduced. Wang et al. \cite{wang2018vehicle} combines the location information of traffic lights from a high-precision map. This approach no longer depends on relative location and generates the traffic light map by collecting prior geographical information of traffic lights. The vision information is then fused with INS using EKF to generate high precision localization result. Soatti et al. \cite{soatti2018implicit} jointly localize non-cooperative physical features in the surrounding, and the information on sensed features are fused by V2V by a consensus procedure. This work presents theoretical solution to the localization problem with simulation. Our work also addresses infrastructure-based localization, and proposed method is validated by real-world experiment. Our ix-node is capable of detecting vehicles with its sensor heads and thus independently provides absolute measurements. This method does not need to assume good initialization and prior knowledge of map.

% Figure
\begin{figure*}[b]
\centering
\includegraphics[width=17.5cm]{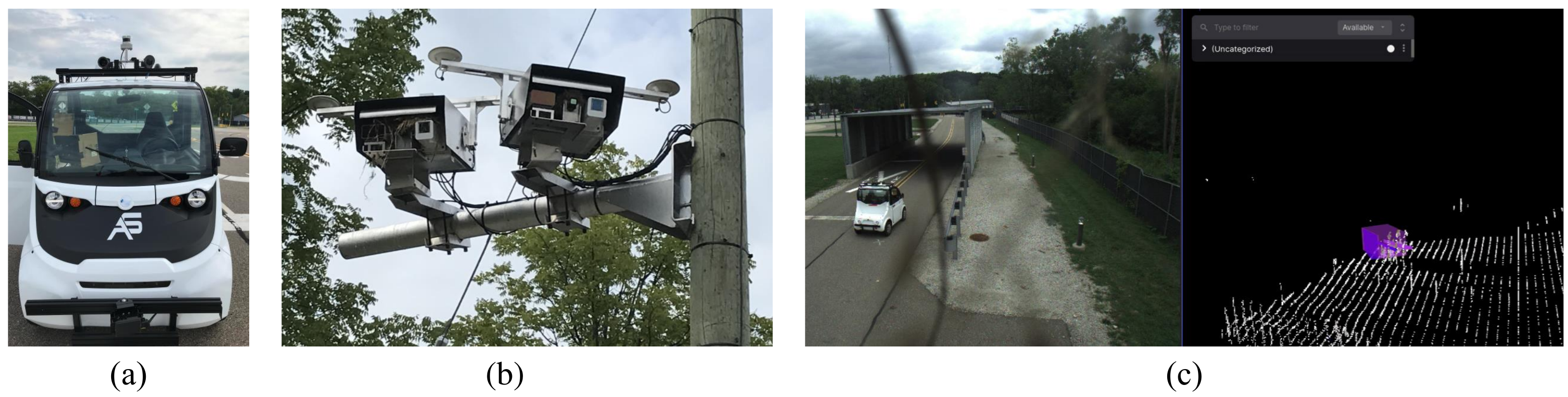}
\caption{
Setup for ix-node perception. (a) Ford's autonomous vehicle. (b) Ix-node with sensor heads. (c) Ix-node's camera image (left) and vehicle candidate generation result as colored marker with LiDAR's point cloud (right).
}
\label{fig:hardware}
\end{figure*}
%

 % Figure - diagram
\begin{figure}[t]
\centering
\includegraphics[height=6.5cm]{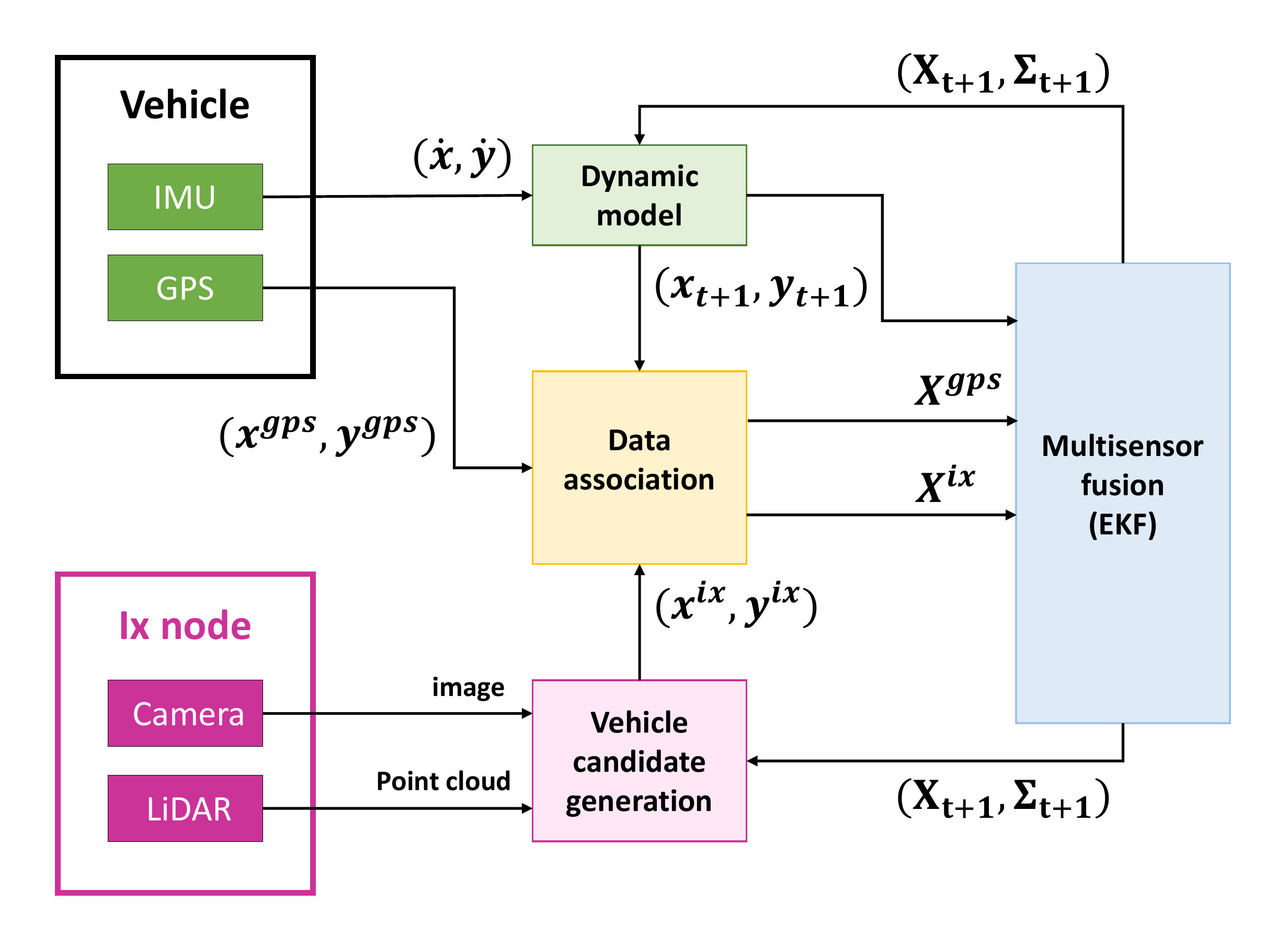}
\caption{
System overview for ix-based vehicle localization.
}
\label{fig:diagram}
\end{figure}
\section{Ix-based Localization \label{sec:ix-based}}

\subsection{System overview}
Let $X_t = [x_t, y_t]^T $ denote the vehicle state vector. This state vector is in East, North, Up (ENU) coordinate frame and relevant to our application since the utilized sensors (e.g. GPS and ix-node) directly measure $x$ and $y$ position of the vehicle. We assume 2D motion and thus resulting output corresponds to East and North components in ENU coordinate frame.

The overall system is summarized in Fig. \ref{fig:diagram}. Vehicle collects linear velocity $(\dot{x},\dot{y})$ using IMU as well as its position $(x^{gps},y^{gps})$ measured by GPS. Ix-node captures image using camera and generates point cloud using LiDAR. The vehicle candidate generation algorithm produces set of tuples $(x^{ix},y^{ix})$ that indicate predicted vehicle positions by ix-node. The vehicle and ix-node data are associated and synchronized in spatial and time domain. Then, EKF-based multisensor fusion is performed to combine both information from the vehicle and ix-node.

%%%%%%%%%%%%%%%%%%%%%%%%%%%%%%%%%%%%%%%%%%%%%%%%%%%%%%%%%%%%%%%%%%%
\subsection{Vehicle candidate generation}
This work utilizes sensors mounted on an infrastructure node, which perceives the vehicle from another perspective. We have deployed multiple infrastructure nodes at each intersection but for the purpose of this study we focus on a single IX node that includes two sensor-heads mounted on a single pole connected to one compute node. Each sensor-head is comprised of a non-spinning LiDAR, a 5MP camera, and infrastructure to power the sensors. The compute node is equipped with an Nvidia TITAN V GPU and a Xeon 2.5GHz CPU and all the processing is done on this compute node. The vehicle and the sensor heads of ix-node are shown in Fig. \ref{fig:hardware}(a) and Fig. \ref{fig:hardware}(b), respectively.

\begin{figure}[t]
\centering
\includegraphics[width=8.5cm]{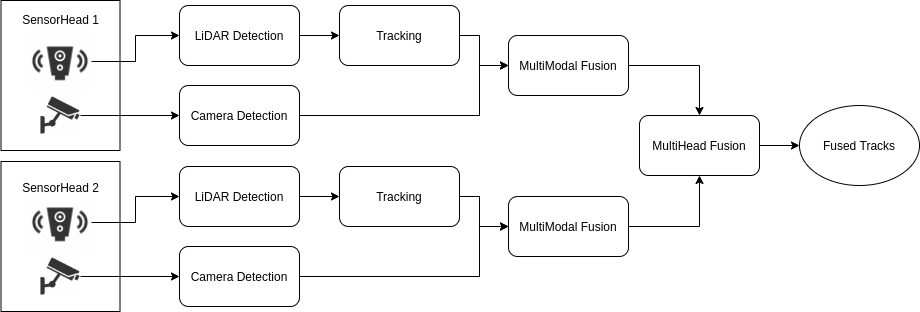}
\caption{
Diagram for ix-node perception algorithm.
}
\label{fig:ix_algorithm}
\end{figure}

The ix-node perceives the nearby objects by the pipeline shown in Fig. \ref{fig:ix_algorithm}. This pipeline implements a late fusion method where detection is run on each sensor modality separately and then the 3D detected objects from LiDAR are tracked and fused with the 2D camera detections to add the classification of objects. Once the tracked objects from each sensor-head are obtained, the tracks from the two sensor-heads are fused to obtain a single view of the environment. We are using YOLO V4 \cite{bochkovskiy2020yolov4} for detecting objects from the images. The 3D detection on LiDAR is done using background removal and then clustering in a Voxel grid representation. For tracking, we're using EKF that includes the center, dimentions, and velocity of the object in the state. The fusion is simply performed by reprojecting the 3D bounding cuboids onto the image and matching the reprojected box with 2D detections using IoU. Finally, the fusion between the two sensor-heads is done using transforming the tracks to UTM coordinates and matching is done using 3D IoU to obtain the final fused tracks from the IX node.

% Figure - diagram

Fig. \ref{fig:hardware}(c) shows the ix-node perception as well as generated vehicle candidates. The left figure shows the camera view of a moving vehicle, and the right figure shows the point cloud by LiDAR sensors and colored markers as detected vehicle candidates.

%%%%%%%%%%%%%%%%%%%%%%%%%%%%%%%%%%%%%%%%%%%%%%%%%%%%%%%%%%%%%%%%%%%
\subsection{Dynamic model (EKF Prediction)}
The motion prediction equation can be written as: $\hat{X}_{t+1} = f(X_t, u_{t+1}) + \alpha_t$ where $\alpha_t$ is the process noise and Gaussian-distributed. We choose a linear constant velocity model for state prediction, expressed as
\begin{align}    
    x_{t+1} &= x_t+\dot{x}_t \Delta t \\
    y_{t+1} &= y_t+\dot{y}_t \Delta t 
\end{align}
where $\dot{x},\dot{y}$ are x and y velocities obtained from IMU, and $\Delta t$ is time interval.\\

The Covariance matrix $\hat{\Sigma}_{t+1}$ of motion prediction is represented by
\begin{align}
\hat{\Sigma}_{t+1} &= G_t \Sigma_t G_t^T + R_{t+1} 
\end{align}
where $G_t$ is the Jacobian matrix of $f(X_t, u_{t+1})$ with respect to $X_t$, $\Sigma_t$ is the covariance matrix of estimation at time $t$, and $R_{t+1}$ is the covariance matrix of the Guassian noise that affects the vehicle state.

%%%%%%%%%%%%%%%%%%%%%%%%%%%%%%%%%%%%%%%%%%%%%%%%%%%%%%%%%%%%%%%%%%%
\subsection{Data association}
Since all the sensors have different timestamps and coordinate systems, we need to synchronize time and associate the data in the same spatial domain. As shown in Fig. \ref{fig:diagram}, we have GPS measurement $(x^{gps},y^{gps})$ and vehicle candidate from ix-node $(x^{ix},y^{ix})$. We synchronize timestamps with respect to POSIX time of the GPS since it has the lowest frequency (10Hz), transform sensor measurements to ENU coordinate frame, and offset the East and North positions by map origin $(x_o,y_o)$. We assume the vehicle has negligible acceleration so the GPS and ix-node outputs are linearized to have the same timestamps. 

For initial filtering of generated vehicle candidates, we set a threshold $d_{thresh}$ to remove the candidates that are too far from the predicted state $(x_{t+1},y_{t+1})$ and also filter out any noise that is out of ix-node detection range. For multiple candidates that are overlapped to each other, we choose a nearest neighbor from the predicted state. Further, the GPS measurements $(x^{gps},y^{gps})$ that is too far from the predicted state $(x_{t+1},y_{t+1})$ are ignored. Filtered GPS and ix-node outputs are fed into EKF-based multisensor fusion module as $X^{gps}$ and $X^{ix}$, respectively.

%%%%%%%%%%%%%%%%%%%%%%%%%%%%%%%%%%%%%%%%%%%%%%%%%%%%%%%%%%%%%%%%%%%
\subsection{Multisensor fusion (EKF measurement update)}

In the system architecture, GPS and ix-node perception output are used as vehicle position measurement. The GPS mounted on the vehicle measures vehicle state in ENU coordinate frame, and the observation equation can be expressed as
\begin{align}
X^{gps}_{t+1} = 
\begin{bmatrix}
x^{gps}_{t+1}\\ y^{gps}_{t+1} 
\end{bmatrix}
=
\begin{bmatrix}
1 & 0\\ 0 & 1
\end{bmatrix}
\begin{bmatrix}
x_{t+1} \\ y_{t+1}
\end{bmatrix}
+ \beta_{gps}
\end{align}
where $[x^{gps}_{t+1}, y^{gps}_{t+1}]^T$ is the GPS measurement and $\beta_{gps}$ is the measurement noise. The covariance matrix of GPS observation error is estimated by
\begin{align}
    Q^{gps}_{t+1} = 
    \begin{bmatrix}
    \delta^2_{x,gps} & \delta_{x,gps}\cdot\delta_{y,gps}\\
    \delta_{x,gps}\cdot\delta_{y,gps} & \delta^2_{y,gps}
    \end{bmatrix}
\end{align}
where $\delta_{x,gps}$ and $\delta_{y,gps}$ are the standard deviations of the GPS measurement noise in $x$ and $y$ directions. 

As for ix-node perception, sensor heads including camera and LiDAR run vehicle tracking algorithms that estimate the vehicle state, and the observation equation is given by 
\begin{align}
X^{ix}_{t+1} = 
\begin{bmatrix}
x^{ix}_{t+1}\\ y^{ix}_{t+1} 
\end{bmatrix}
=
\begin{bmatrix}
1 & 0\\ 0 & 1
\end{bmatrix}
\begin{bmatrix}
x_{t+1} \\ y_{t+1}
\end{bmatrix}
+ \gamma_{ix}
\end{align}
where $[x^{ix}_{t+1}, y^{ix}_{t+1}]^T$ is the ix-node perception output and $\gamma_{ix}$ is the measurement noise. Similar to the case of GPS, the covariance matrix of ix-node perception error is given by
\begin{align}
    Q^{ix}_{t+1} = 
    \begin{bmatrix}
    \delta^2_{x,ix} & \delta_{x,ix}\cdot\delta_{y,ix}\\
    \delta_{x,ix}\cdot\delta_{y,ix} & \delta^2_{y,ix}
    \end{bmatrix}
\end{align}
where $\delta_{x,ix}$ and $\delta_{y,ix}$ are the standard deviations of the ix-node perception algorithm noise in $x$ and $y$ directions. We refer to Najjar et al. \cite{el2005road} for detailed derivation of the covariance matrix. 

%%%%%%%%%%%%%%%%%%%%%%%%%%%%%%%%%%%%%%%%%%%%%%%%%%%%%%%%%%%%%%%%%%%
%\subsection{Multisensor fusion (Measurement update)}
We first estimate the vehicle position using GPS measurements and then incorporate ix-node perception outputs. The innovation $\nu^{gps}_{t+1}$ of the GPS measurement is 
\begin{align}
    \nu^{gps}_{t+1} = x^{gps}_{t+1} - 
    \begin{bmatrix}
    x_{t+1}\\
    y_{t+1}
    \end{bmatrix}
\end{align}
The covariance matrix of innovation is
\begin{align}
    S^{gps}_{t+1} = H^{gps}_{t+1} \hat{\Sigma}_{t+1}(H^{gps}_{t+1})^T + Q^{gps}_{t+1}
\end{align}
where $H^{gps}_{t+1}$ is the Jacobian matrix of $\hat{X}_{t+1}$ with respect to $X^{gps}_{t+1}$. Kalman gain, corrected position, and associated covariance matrix of GPS measurement are
\begin{align}
    K^{gps}_{t+1} &= \hat{\Sigma}_{t+1} (H^{gps}_{t+1})^T (S^{gps}_{t+1})^{-1}\\
    \overline{X}_{t+1} &= 
    \begin{bmatrix}
    x_{t+1}\\
    y_{t+1}
    \end{bmatrix}
    + K^{gps}_{t+1} \nu^{gps}_{t+1}\\
    \overline{\Sigma}_{t+1} &= (I - K^{gps}_{t+1} H^{gps}_{t+1}) \hat{\Sigma}_{t+1}
\end{align}

Next, the innovation $\nu^{ix}_{t+1}$ of the ix-perception output is 
\begin{align}
    \nu^{ix}_{t+1} = x^{ix}_{t+1} - \overline{X}_{t+1}
\end{align}
The covariance matrix of innovation is
\begin{align}
    S^{ix}_{t+1} = H^{ix}_{t+1} \overline{\Sigma}_{t+1}(H^{ix}_{t+1})^T + Q^{ix}_{t+1}
\end{align}
where $H^{ix}_{t+1}$ is the Jacobian matrix of $\overline{X}_{t+1}$ with respect to $X^{ix}_{t+1}$. Kalman gain of ix-node measurement is
\begin{align}
    K^{ix}_{t+1} = \overline{\Sigma}_{t+1} (H^{ix}_{t+1})^T (S^{ix}_{t+1})^{-1}
\end{align}
The final output from Extended Kalman Filter for vehicle position is given as   
\begin{align}
X_{t+1} = \overline{X}_{t+1} + K^{ix}_{t+1} \nu^{ix}_{t+1}    
\end{align}
with associated covariance matrix
\begin{align}
    \Sigma_{t+1} = (I - K^{ix}_{t+1} H^{ix}_{t+1}) \overline{\Sigma}_{t+1}
\end{align}

\section{Experiments \label{sec:experiments}}

%%%%%%%%%%%%%%%%%%%%%%%%%%%%%%%%%%%%%%%%%%%%%%%%%%%%%%%%%%%%%%%%%%%
\subsection{Experimental setup}
\begin{figure}[t]
\centering
\includegraphics[width=.47\textwidth]{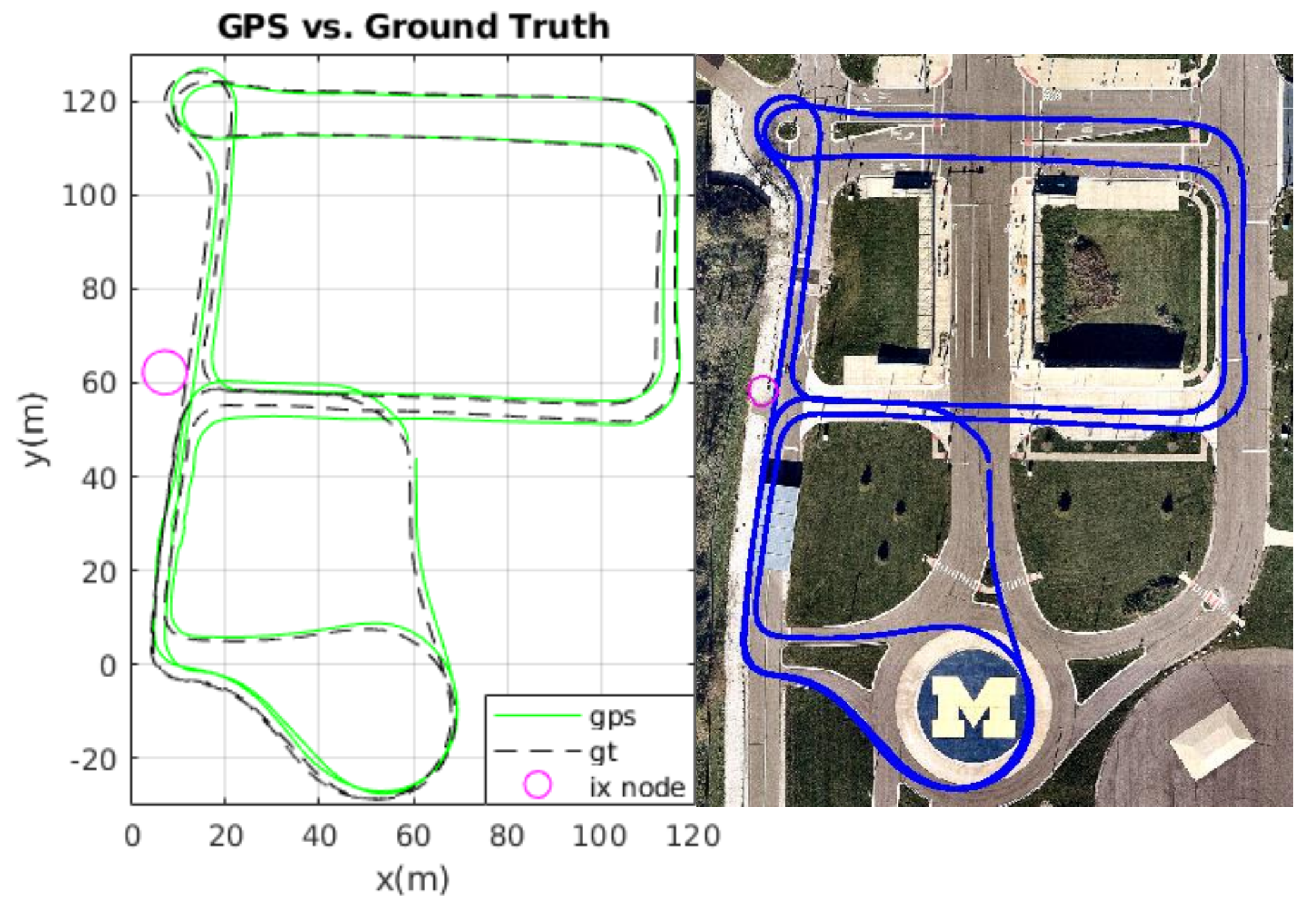}
\caption{
Vehicle's trajectories measured by GPS and ground truth device with ix-node position (left) and corresponding satellite view map with the ground truth marked in blue (right).
}
\label{fig:map}
\end{figure}

To validate our algorithm, we run Ford's autonomous vehicle (Fig. \ref{fig:hardware}(a)) at Mcity test track \cite{mcity}, a state-of-the-art testing facility designed for testing performance of autonomous vehicles in Ann Arbor, Michigan. The vehicle carries a GPS unit and a Novatel ProPak6 that provides a ground truth trajectory by fusing it's RTK corrected GPS and IMU measurements. Fig. \ref{fig:map} shows the entire trajectory of the vehicle that is run for 300 seconds on the test track. The ground truth Novatel system is mounted on the vehicle at the position that is distance $d_{GNSS}$ ahead of the GPS. Then, the ground truth measurement $X^{gt}$ in ENU coordinate frame can be expressed as
\begin{align}
    X^{gt} = 
    \begin{bmatrix}
        x^{gt}-d\cdot \cos(\theta+\frac{\pi}{2})\\
        y^{gt}-d\cdot \sin(\theta+\frac{\pi}{2})
    \end{bmatrix}
\end{align}
where $(x^{gt},x^{gt})$ are raw measurement for vehicle position and $\theta$ is vehicle orientation measured by Novatel's device. Fig. \ref{fig:map} shows ix-node position $(I_x,I_y)$, overall trajectories for GPS and ground truth, and satellite view map of Mcity.

% Figure
%

% Figure
\begin{figure}[b]
\centering
\includegraphics[width=.52\textwidth]{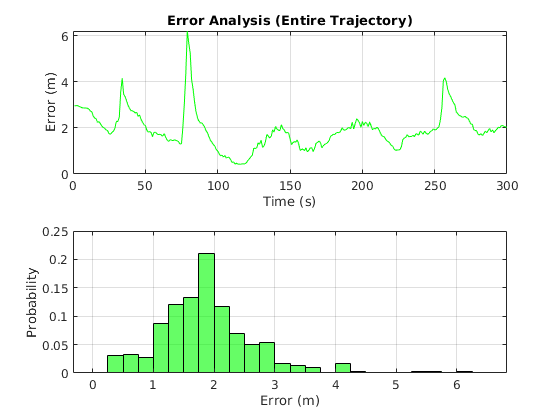}
\caption{
Error analysis of entire trajectory. The top shows the localization error plot of GPS measurement for vehicle's 300 seconds run at Mcity, and the bottom shows the corresponding localization error histogram.
}
\label{fig:error_entire_traj}
\end{figure}

% Figure
\begin{figure*}[t!]
  \centering
    \includegraphics[width=18cm]{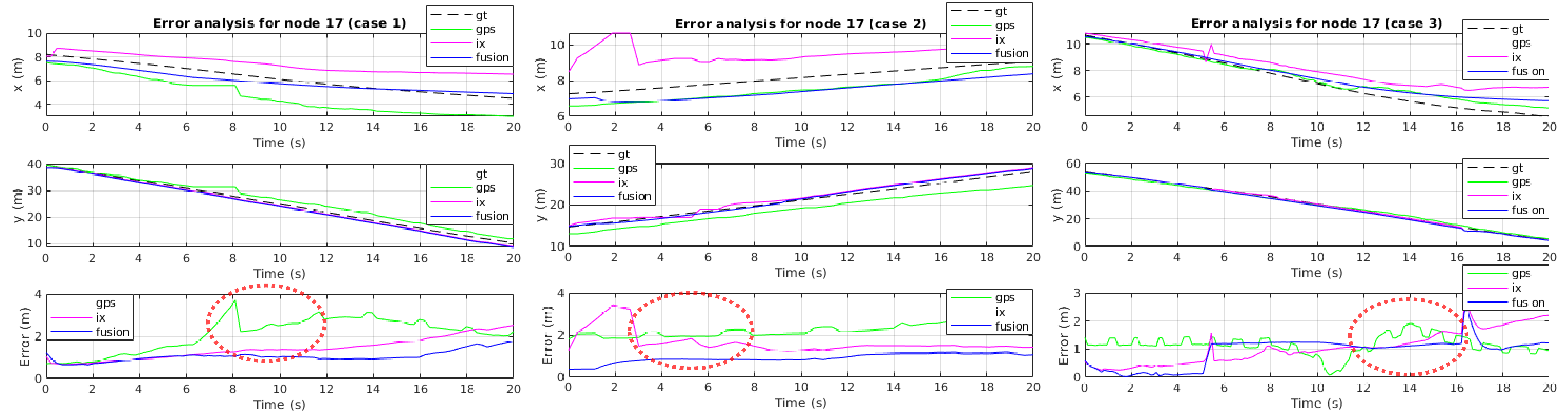}
  \caption{Error analysis of ix-based localization method for three cases when vehicle is running nearby ix-node. Longitudinal and lateral position are provided by ground truth, GPS, ix-node, and proposed algorithm. The localization RMSE is estimated by GPS, ix-node, and proposed fusion method. The circled regions in red indicate the intervals when the vehicle drives in a tunnel.}
  \label{fig:error_plots}
\end{figure*}

% Figure
\begin{figure*}[t!]
  \centering
    \includegraphics[width=18cm]{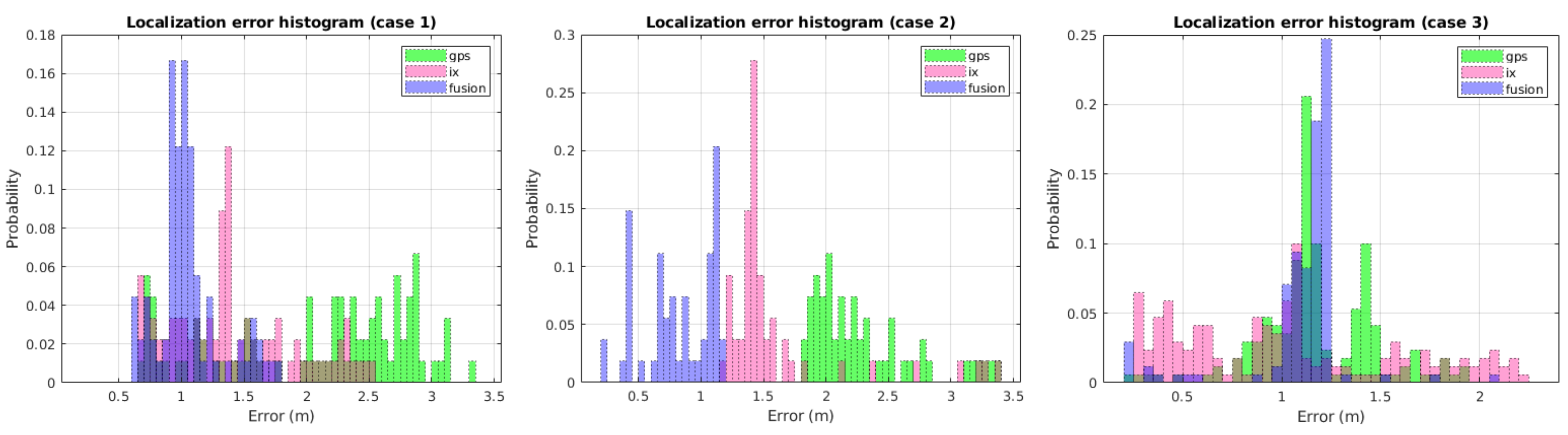}
  \caption{Localization error histograms for GPS, ix, and fusion outputs. In all cases, fusion improves localization performance from using GPS data.}
  \label{fig:histogram}
\end{figure*}

%%%%%%%%%%%%%%%%%%%%     Table 1     %%%%%%%%%%%%%%%%%%%%%     
\begin{table}[t]
\centering
\caption{Position Error for GPS (Entire Trajectory)}
\label{tab1}
\begin{center}
\begin{tabular}{c | c | c}
\hline
Error & Mean & Std. \\
\hline
Longitudinal error (m) & 1.17 & 0.65 \\
\hline
Lateral error (m) & 1.20 & 1.05 \\
\hline
Total error (m) & 1.91 & 0.82 \\
\hline
\end{tabular}
\end{center}
\end{table}

Fig. \ref{fig:error_entire_traj} shows the error analysis for entire trajectory from Fig. \ref{fig:map}. The top shows the root mean square error (RMSE) for estimated trajectory, and the bottom shows localization error histogram. It can be seen that the raw GPS error is non-trivial (e.g. the maximum error is greater than 5m), so the fusion with ix-node is necessary and improves the localization performance. Table \ref{tab1} summarizes statistics of GPS error for entire trajectory.

%%%%%%%%%%%%%%%%%%%%%%%%%%%%%%%%%%%%%%%%%%%%%%%%%%%%%%%%%%%%%%%%%%%
\subsection{Multisensor fusion for ix-based localization}
To evaluate ix-based localization performance, we study three cases where vehicle drives nearby an ix-node. The camera image of Fig. \ref{fig:hardware}(c) shows an instance of such case. For each case, the vehicle is within ix-node's detection range so we can execute multisensor fusion algorithm combining GPS, IMU, and ix-node perception.

%%%%%%%%%%%%%%%%%%%%     Table 2     %%%%%%%%%%%%%%%%%%%%%     
\begin{table}[b]
\centering
\caption{Parameters for Ix-based Localization}
\label{tab2}
\begin{center}
\begin{tabular}{c | c | c}
\hline
Symbol & Description & Value\\
\hline
$\Delta t$ & Time interval for dynamic model & 0.1s \\
$x_o$ & East position of map origin & 277495m \\
$y_o$ & North position of map origin & 4686600m \\
$d_{thresh}$ & Threshold for filtering vehicle candidate & 5m \\
$\delta_{x,gps}$ & Longitudinal std. deviation of GPS noise & 0.8m \\
$\delta_{y,gps}$ &  Lateral std. deviation of GPS noise & 2m \\
$\delta_{x,ix}$ & Longitudinal std. deviation of ix-node noise & $f(X^{ix})$ \\
$\delta_{y,ix}$ & Lateral std. deviation of ix-node noise & 0.3m \\
$d_{GNSS}$ & Distance of ground truth system from GPS & 0.0381m \\
$I_x$ & Longitudinal ix-node position w.r.t map origin & 7.13m \\
$I_y$ & Lateral ix-node position w.r.t map origin & 62.19m \\

\hline
\end{tabular}
\end{center}
\end{table}

Table \ref{tab2} shows the details of used parameters for ix-based localization. For the longitudinal standard deviation of ix-node perception noise $\delta_{x,ix}$, we define the value as
\begin{align}
    f(X^{ix}) = |0.051\cdot d_{ix} - 0.702|
\end{align}
where $d_{ix} = \sqrt{(x^{ix}-I_x)^2+(x^{ix}-I_y)^2}$. We have observed that the ix-node perception error increases as the object being detected gets farther away from the ix-node; thus, we model the non-linear covariance matrix that is dynamically changing based on the distance between ix-node and vehicle.

Fig. \ref{fig:error_plots} shows the multisensor fusion results for the three cases. We observe longitudinal and lateral positions provided by ground truth, GPS, ix-node, and proposed fusion method. The RMSE of GPS, ix-node, and fusion method are also plotted. Circled regions in red indicate the intervals when the vehicle runs in a tunnel. Fig. \ref{fig:histogram} displays corresponding localization error histograms for each case. Quantitative sensor statistics for the position error of each case is detailed in Table \ref{tab3}. These errors measure the performance of our localization approach. 

%%%%%%%%%%%%%%%%%%%%     Table 3     %%%%%%%%%%%%%%%%%%%%%     
\begin{table}[t]
\centering
\caption{Mean and Standard Deviation for Position Error}
\label{tab3}
\begin{center}
\begin{tabular}{c | c | c | c | c | c | c}
\hline
Sensor & GPS  & GPS  & Ix   & Ix   & Fusion & Fusion \\
Statistics & Mean & Std. & Mean & Std. & Mean & Std. \\
\hline
Case 1 (m) & 2.09 & 0.78 & 1.40 & 0.50 & \textbf{1.05} & \textbf{0.25} \\
\hline
Case 2 (m) & 2.30 & 0.43 & 1.62 & 0.53 & \textbf{0.89} & \textbf{0.22} \\
\hline
Case 3 (m) & 1.15 & 0.33 & 1.05 & 0.59 & \textbf{0.92} & \textbf{0.49} \\
\hline
Average (m) & 1.85 & 0.51 & 1.36 & 0.54 & \textbf{0.95} & \textbf{0.32} \\
\hline
\end{tabular}
\end{center}
\end{table}

From all the cases, we observe that the GPS measurement is noisy in both longitudinal and lateral direction, the mean position errors (raw GPS) for the case 1 and 2 exceed two meters. The ix-node perception has lower position error than GPS output in all cases but higher average standard deviation for position errors. The case 2 and 3 from Fig. \ref{fig:error_plots} show that there exist instances where ix-node perception output fails but it recovers quickly. The ix-node perception error comes from vehicle occlusion and sensor noise so the position of generated vehicle candidates may deviate from the ground truth further than the position estimated by GPS that is Gaussian-distributed from the ground truth. By fusing these two sources of measurement, we see that the mean position errors are improved in all cases, achieving an average of 0.95 meter. The localization error histograms in Fig. \ref{fig:histogram} also demonstrate that mean errors get shifted to the left when fusing GPS and ix data in case 1 and 2. Although mean errors in case 3 seem similar for ix and fusion, fused data has less standard deviation. In all cases, the mean errors of fusion are reduced from using GPS.

It is also worth noting that the ix-node perception works even if the vehicle is moving in a tunnel. As shown in Fig. \ref{fig:hardware}(c), our ix-node is built so that its detection range covers the case when a vehicle is running in a tunnel. The circled regions in Fig. \ref{fig:error_plots} shows high peaks of GPS errors inside the tunnel due to GPS attenuation. Ix-node still detects the vehicle with sensor heads, and the fusion result shows that ix-node compensates for GPS failure in a tunnel.

\section{Conclusion \label{sec:conclusion}}
This paper realizes infrastructure node-based vehicle localization. This work utilizes additional sensors mounted on an infrastructure node, namely ix-node, which perceives the vehicle from another perspective. An ix-node has two sensor heads, where each sensor head consists of a camera and a LiDAR. The ix-node perceives the nearby vehicles, and the vehicles can localize themselves better via communication with the ix-node. 

We execute ix-based localization framework with Extended Kalman Filter including data association and time synchronization between the vehicle and ix-node. The ix-node has its own perception algorithm to generate vehicle candidates, and the outputs from dynamic model are associated with the generated candidates to be fed into the multifusion system. We test the proposed algorithm using Ford autonomous vehicle and ix-node built on a test track at Mcity in Ann Arbor, Michigan. The result shows that the proposed ix-based localization performs better than GPS and ix perception in all cases, achieving less than one meter of mean position error. It is also observed that the ix-node compensates for GPS failure in a tunnel. 

The future work will implement ix-based localization framework with multiple ix-node. We intend to explore combining information at the sensor level like the raw images and pointclouds instead of the perception output. 

%%%%%%%%%%%%%%%%%%%%%%%%%%%%%%%%%%%%%%%%%%%%%%%%%%%%%%%%%%%%%%%%%%%%%%%%%%%%%%%%

% \addtolength{\textheight}{-12cm}   % This command serves to balance the column lengths
                                  % on the last page of the document manually. It shortens
                                  % the textheight of the last page by a suitable amount.
                                  % This command does not take effect until the next page
                                  % so it should come on the page before the last. Make
                                  % sure that you do not shorten the textheight too much.

%%%%%%%%%%%%%%%%%%%%%%%%%%%%%%%%%%%%%%%%%%%%%%%%%%%%%%%%%%%%%%%%%%%%%%%%%%%%%%%%
%\section*{APPENDIX \label{sec:appendix}}
%\subsection*{Constants and functions in \ref{sec:robustness}}
%\noindent The constants for bounding the cross terms are:

% \section*{ACKNOWLEDGMENT}
% \noindent  The authors would  like to acknowledge valuable discussions with XXX.

%%%%%%%%%%%%%%%%%%%%%%%%%%%%%%%%%%%%%%%%%%%%%%%%%%%%%%%%%%%%%%%%%%%%%%%%%%%%%%%%

%
%\begin{thebibliography}{10}
%
%\bibitem{Zarchan1997}
%P.~Zarchan.
%\newblock {Tactical and strategic missile guidance}.
%\newblock {\em Progress in astronautics and aeronautics}, 176, 2002.
%
%\end{thebibliography}

%\bibliography{../../../../Dropbox/papers/bibs/CDC2018}
%\bibliographystyle{unsrt}

%%%{\footnotesize
\bibliography{reference}

\begin{thebibliography}{10}

\bibitem{reid2019localization}
Tyler~GR Reid, Sarah~E Houts, Robert Cammarata, Graham Mills, Siddharth
  Agarwal, Ankit Vora, and Gaurav Pandey.
\newblock Localization requirements for autonomous vehicles.
\newblock {\em arXiv preprint arXiv:1906.01061}, 2019.

\bibitem{ixpatent}
Linjun Zhang, Helen~Elizabeth Kourous-Harrigan, Ankit~Girish Vora, Cordin
  Cionca, Lu~Xu, and Jeffrey~Thomas Remillard.
\newblock Transportation infrastructure communication and control, June~6 2018.
\newblock US Patent App. 16/018,144.

\bibitem{ixmedium}
Tony Lockwood.
\newblock Exploring how smart infrastructure can help ford build a great self
  driving service for miamians, 2021.

\bibitem{challita2009application}
Georges Challita, St{\'e}phane Mousset, Fawzi Nashashibi, and Abdelaziz
  Bensrhair.
\newblock An application of v2v communications: Cooperation of vehicles for a
  better car tracking using gps and vision systems.
\newblock In {\em 2009 IEEE Vehicular Networking Conference (VNC)}, pages 1--6.
  IEEE, 2009.

\bibitem{rife2011collaborative}
Jason Rife.
\newblock Collaborative vision-integrated pseudorange error removal:
  Team-estimated differential gnss corrections with no stationary reference
  receiver.
\newblock {\em IEEE Transactions on Intelligent Transportation Systems},
  13(1):15--24, 2011.

\bibitem{bento2012inter}
Luis~Conde Bento, Ricardo Parafita, and Urbano Nunes.
\newblock Inter-vehicle sensor fusion for accurate vehicle localization
  supported by v2v and v2i communications.
\newblock In {\em 2012 15th International IEEE Conference on Intelligent
  Transportation Systems}, pages 907--914. IEEE, 2012.

\bibitem{bonnifait2001data}
Philippe Bonnifait, Pascal Bouron, Paul Crubille, and Dominique Meizel.
\newblock Data fusion of four abs sensors and gps for an enhanced localization
  of car-like vehicles.
\newblock In {\em Proceedings 2001 ICRA. IEEE International Conference on
  Robotics and Automation (Cat. No. 01CH37164)}, volume~2, pages 1597--1602.
  IEEE, 2001.

\bibitem{el2005road}
Maan~E El~Najjar and Philippe Bonnifait.
\newblock A road-matching method for precise vehicle localization using belief
  theory and kalman filtering.
\newblock {\em Autonomous Robots}, 19(2):173--191, 2005.

\bibitem{levinson2010robust}
Jesse Levinson and Sebastian Thrun.
\newblock Robust vehicle localization in urban environments using probabilistic
  maps.
\newblock In {\em 2010 IEEE international conference on robotics and
  automation}, pages 4372--4378. IEEE, 2010.

\bibitem{lee2017feature}
Elijah~S Lee and Dongsuk Kum.
\newblock Feature-based lateral position estimation of surrounding vehicles
  using stereo vision.
\newblock In {\em 2017 IEEE Intelligent Vehicles Symposium (IV)}, pages
  779--784. IEEE, 2017.

\bibitem{lee2019bird}
Elijah~S Lee, Wongun Choi, and Dongsuk Kum.
\newblock Bird’s eye view localization of surrounding vehicles: Longitudinal
  and lateral distance estimation with partial appearance.
\newblock {\em Robotics and Autonomous Systems}, 112:178--189, 2019.

\bibitem{localizationtechniquepatent}
Siddharth Agarwal, Ankit~Girish Vora, Jakob~Nikolaus Hoellerbauer, and Peng-Yu
  Chen.
\newblock Localization technique selection, February~2 2018.
\newblock US Patent App. 15/907,623.

\bibitem{vora2020aerial}
Ankit Vora, Siddharth Agarwal, Gaurav Pandey, and James McBride.
\newblock Aerial imagery based lidar localization for autonomous vehicles.
\newblock {\em arXiv preprint arXiv:2003.11192}, 2020.

\bibitem{wei2011intelligent}
Lijun Wei, Cindy Cappelle, Yassine Ruichek, and Fr{\'e}d{\'e}rick Zann.
\newblock Intelligent vehicle localization in urban environments using
  ekf-based visual odometry and gps fusion.
\newblock {\em IFAC Proceedings Volumes}, 44(1):13776--13781, 2011.

\bibitem{kummerle2009autonomous}
Rainer Kummerle, Dirk Hahnel, Dmitri Dolgov, Sebastian Thrun, and Wolfram
  Burgard.
\newblock Autonomous driving in a multi-level parking structure.
\newblock In {\em 2009 IEEE International Conference on Robotics and
  Automation}, pages 3395--3400. IEEE, 2009.

\bibitem{wan2018robust}
Guowei Wan, Xiaolong Yang, Renlan Cai, Hao Li, Yao Zhou, Hao Wang, and Shiyu
  Song.
\newblock Robust and precise vehicle localization based on multi-sensor fusion
  in diverse city scenes.
\newblock In {\em 2018 IEEE International Conference on Robotics and Automation
  (ICRA)}, pages 4670--4677. IEEE, 2018.

\bibitem{wolcott2017robust}
Ryan~W Wolcott and Ryan~M Eustice.
\newblock Robust lidar localization using multiresolution gaussian mixture maps
  for autonomous driving.
\newblock {\em The International Journal of Robotics Research}, 36(3):292--319,
  2017.

\bibitem{schaupp2019oreos}
Lukas Schaupp, Mathias B{\"u}rki, Renaud Dub{\'e}, Roland Siegwart, and Cesar
  Cadena.
\newblock Oreos: Oriented recognition of 3d point clouds in outdoor scenarios.
\newblock In {\em 2019 IEEE/RSJ International Conference on Intelligent Robots
  and Systems (IROS)}, pages 3255--3261. IEEE, 2019.

\bibitem{lu2019deepvcp}
Weixin Lu, Guowei Wan, Yao Zhou, Xiangyu Fu, Pengfei Yuan, and Shiyu Song.
\newblock Deepvcp: An end-to-end deep neural network for point cloud
  registration.
\newblock In {\em Proceedings of the IEEE/CVF International Conference on
  Computer Vision}, pages 12--21, 2019.

\bibitem{li2018map}
Franck Li, Philippe Bonnifait, and Javier Iba{\~n}ez-Guzm{\'a}n.
\newblock Map-aided dead-reckoning with lane-level maps and integrity
  monitoring.
\newblock {\em IEEE Transactions on Intelligent Vehicles}, 3(1):81--91, 2018.

\bibitem{asghar2020vehicle}
Rabbia Asghar, Mario Garz{\'o}n, Jerome Lussereau, and Christian Laugier.
\newblock Vehicle localization based on visual lane marking and topological map
  matching.
\newblock In {\em 2020 IEEE International Conference on Robotics and Automation
  (ICRA)}, pages 258--264. IEEE, 2020.

\bibitem{wang2018vehicle}
Chunxiang Wang, Hairu Huang, Yang Ji, Bing Wang, and Ming Yang.
\newblock Vehicle localization at an intersection using a traffic light map.
\newblock {\em IEEE Transactions on Intelligent Transportation Systems},
  20(4):1432--1441, 2018.

\bibitem{soatti2018implicit}
Gloria Soatti, Monica Nicoli, Nil Garcia, Benoit Denis, Ronald Raulefs, and
  Henk Wymeersch.
\newblock Implicit cooperative positioning in vehicular networks.
\newblock {\em IEEE Transactions on Intelligent Transportation Systems},
  19(12):3964--3980, 2018.

\bibitem{bochkovskiy2020yolov4}
Alexey Bochkovskiy, Chien-Yao Wang, and Hong-Yuan~Mark Liao.
\newblock Yolov4: Optimal speed and accuracy of object detection.
\newblock {\em arXiv preprint arXiv:2004.10934}, 2020.

\bibitem{mcity}
University of~Michigan.
\newblock Mcity test facility, 2016.

\end{thebibliography}
\bibliographystyle{unsrt}

\end{document}